\documentclass{article}

\usepackage[preprint]{css}

\usepackage[utf8]{inputenc} 
\usepackage[T1]{fontenc}    
\usepackage{hyperref}       
\usepackage{url}            
\usepackage{booktabs}       
\usepackage{amsfonts}       
\usepackage{nicefrac}       
\usepackage{microtype}      
\usepackage{graphicx}
\usepackage[ruled,vlined]{algorithm2e}
\usepackage{algorithmic}
\usepackage{breakurl}
\title{TalentMine: LLM-Based Extraction and Question-Answering from Multimodal Talent Tables}

\author{%
  Varun Mannam\thanks{Corresponding author: Dr. Varun Mannam is an Applied Scientist II in People eXperience Technology (PXT) Central Science at Amazon, where he leads research initiatives in talent analytics and intelligent HR systems.} \\
  PXT Central Science \\
  Amazon\\
  Seattle, WA 98004 \\
  \texttt{mannamvs@amazon.com} \\
  \And
  Fang Wang \\ 
  PXT Central Science \\
  Amazon\\
  Seattle, WA 98004 \\
  \texttt{fwfang@amazon.com} \\
  \AND
  Chaochun Liu \\ 
  PXT Central Science \\
  Amazon\\
  Seattle, WA 98004 \\
  \texttt{chchliu@amazon.com} \\
  \And
  Xin Chen \\ 
  PXT Central Science \\
  Amazon\\
  Seattle, WA 98004 \\
  \texttt{xcaa@amazon.com} \\
}
\date{TODAY}

\newcommand{\keywords}[1]{\par\noindent{\small\textbf{Keywords:} #1}}
\begin{document}
\vspace{-1cm}
\maketitle
\vspace{-1cm}
\begin{abstract}
In talent management systems, critical information often resides in complex tabular formats, presenting significant retrieval challenges for conventional language models. These challenges are pronounced when processing Talent documentation that requires precise interpretation of tabular relationships for accurate information retrieval and downstream decision-making. Current table extraction methods struggle with semantic understanding, resulting in poor performance when integrated into retrieval-augmented chat applications. This paper identifies a key bottleneck - while structural table information can be extracted, the semantic relationships between tabular elements are lost, causing downstream query failures. To address this, we introduce TalentMine, a novel LLM-enhanced framework that transforms extracted tables into semantically enriched representations. Unlike conventional approaches relying on CSV or text linearization, our method employs specialized multimodal reasoning to preserve both structural and semantic dimensions of tabular data. Experimental evaluation across employee benefits document collections demonstrates TalentMine's superior performance, achieving 100\% accuracy in query answering tasks compared to 0\% for standard AWS Textract extraction and 40\% for AWS Textract Visual Q\&A capabilities. Our comparative analysis also reveals that the Claude v3 Haiku model achieves optimal performance for talent management applications. The key contributions of this work include (1) a systematic analysis of semantic information loss in current table extraction pipelines, (2) a novel LLM-based method for semantically enriched table representation, (3) an efficient integration framework for retrieval-augmented systems as end-to-end systems, and (4) comprehensive benchmarks on talent analytics tasks showing substantial improvements across multiple categories.
\end{abstract}
\vspace{-0.2cm}
\vspace{-0.5cm}

\section{Introduction}
\begin{figure*}[!t]
  \centering
 \includegraphics[width=1\linewidth]{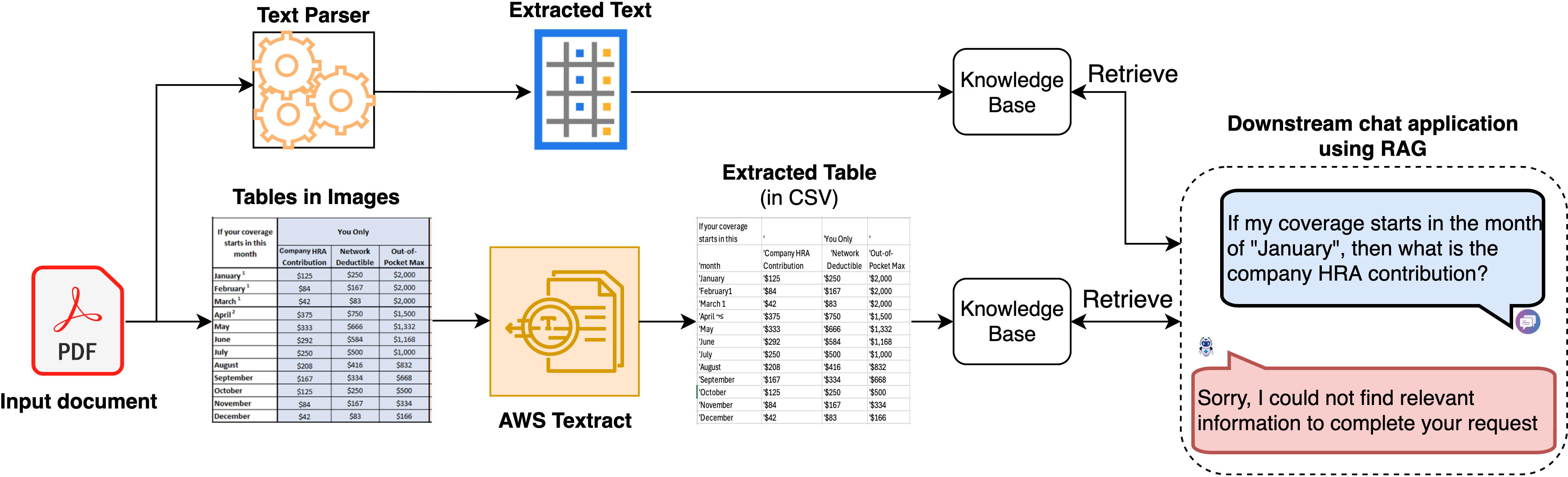}
  \caption{Existing workflow showing PDF document processing using AWS Textract to extract tabular data, which is then indexed for contextual retrieval by a RAG-powered chat system responding to structured data queries. \textit{Note: All monetary values have been masked and replaced with simulated data.}}
  \label{current-system}
\end{figure*}

In today's dynamic talent management landscape, organizations face increasing challenges in processing and analyzing vast amounts of HR-related documents containing critical employee information. Fortune 500 companies with complex HR structures report significant inefficiencies in their existing document processing systems, seeking at least a two-fold improvement in accuracy for processing critical HR documents such as benefits information, compensation details, and performance metrics. While optical character recognition (OCR) technologies \cite{amazon_textract, hegghammer2022ocr} have advanced significantly, extracting structured information from employee benefits documents remains challenging due to their diverse formats, complex layouts, and the need for maintaining data accuracy and compliance \cite{zhong2019publaynet, kavasidis2022chartocr}. Talent management systems specifically highlighted the need for robust solutions that can handle complex organizational hierarchies and multi-tiered benefit structures and documents. Recent developments in deep learning, particularly in convolutional neural networks (CNNs) \cite{zhong2019publaynet} and large language models (LLMs) \cite{qian2019exploring}, have opened new possibilities for intelligent talent management systems \cite{li2022structurnet, huang2022cycleformer}. These advancements are particularly relevant for HR professionals who need to efficiently process various documents, including employee evaluations, training records, and organizational charts. Traditional OCR models, while useful, often struggle with accurately recognizing text in structured and/or unstructured formats common in HR documentation \cite{zhong2020image, li2020tablebank, shahab2010open}, potentially impacting critical talent management decisions.

The integration of modern AI techniques in HR document processing has become crucial for several key talent management applications \cite{gobel2013icdar, gao2019icdar, li2019tablebank}. For instance, Retrieval Augmented Generation (RAG) systems are increasingly vital for HR knowledge management, enabling quick access to relevant employee information and policy documents. Similarly, automated query systems (comparable to Text-to-SQL) allow HR professionals to naturally interact with employee databases \cite{amazon_textract}, facilitating data-driven decision-making in areas such as talent acquisition, retention, and development.

Our research addresses these challenges by developing an intelligent system that combines advanced OCR capabilities with LLMs, specifically designed for talent management computing. This system not only extracts information accurately from multimodal HR documents but also enables sophisticated analysis and query capabilities, supporting various organizational management tasks. The approach is particularly relevant for modern HR practices that require quick, accurate, and compliant processing of employee-related documentation while maintaining data integrity and supporting fair talent management practices. The current HR document processing systems face significant challenges in extracting and understanding the relevant information needed to address complex queries. To illustrate this, consider a scenario where an HR professional needs to retrieve benefits information from employee documents. A typical application, such as the one shown in Figure~\ref{current-system}, utilizes Amazon's AI services, including Amazon Textract to extract structured data from documents, Amazon Textract with visual Q\& A to extract structured data from documents, including user's question-answering in the offline mode, and Amazon Q, a question-answering service, to enable querying of that data. While traditional systems like Amazon Textract can extract tabular data from input documents and convert them into structured formats, the extracted information often lacks the necessary contextual understanding and contextual relevance required to provide accurate responses to complex queries. For example, when handling a query such as \textit{If my coverage starts in the month of "January", then what is the company HRA contribution?}, the application may respond with an error message, stating, \textit{Sorry, I could not find relevant information to complete your request.} This highlights a critical challenge in current table extraction methods - while the structural information is preserved in CSV format, the system struggles to retrieve the requested January contribution amount from the extracted table data, even when the information is present. In summary, the current HR document processing systems face limitations in their ability to extract, understand, and retrieve the relevant information needed to address complex, context-specific queries, resulting in incomplete or irrelevant responses.

Our research addresses these limitations by introducing  a paradigm-shifting innovative LLM-based approach specifically designed for talent management applications. This paper makes the following key contributions:
\begin{itemize}
    \item We present a novel LLM-based method for extracting and interpreting HR-related table information, offering superior flexibility and accuracy compared to traditional approaches. This advancement particularly benefits talent acquisition, performance management, and benefits administration processes.
    \item We identify critical limitations in existing HR document processing systems that impact talent management effectiveness, particularly in handling complex organizational documents and employee benefit information, which are both structured and unstructured data.
    \item We demonstrate significant improvements in recall and reliability of HR document processing, achieving perfect recall in our evaluation scenarios, thereby enabling more effective talent management decision-making and employee service delivery.
\end{itemize}
Our approach, while initially focused on HR document processing, establishes a framework that can be extended to various organizational applications beyond HR, including career development tracking, performance evaluation systems, and organizational planning tools. This research advances the field of enterprise computing by providing more robust and efficient data processing capabilities, ultimately supporting organizations in their quest to better manage and develop their operational systems and human capital. The remainder of this paper is organized as follows: Section~\ref{sec2} reviews related work in organizational computing and document processing. Section~\ref{sec3} details our LLM-based methodology. Section~\ref{sec4} presents experimental results and real-world applications in HR scenarios. Finally, Section~\ref{sec5} discusses implications for enterprise management practices and future research directions.

\section{Related Work} \label{sec2}
The evolution of talent management systems has highlighted the critical need for efficient processing of HR documentation, particularly in extracting structured information from employee-related documents. Traditionally, HR departments relied heavily on manual data entry and verification, a process that was not only time-consuming but also prone to errors, especially when dealing with complex employee documents containing tables of benefits, compensation, or performance metrics. These challenges are particularly pronounced when processing tables embedded in images within PDFs, which contain sensitive information like pay ranges, benefits contributions, and performance criteria. While solutions like Amazon Mechanical Turk (MTurk) \cite{mtruck} exist for crowdsourcing such tasks, they are often unsuitable for HR applications due to data privacy concerns and the confidential nature of employee information.

Recent advances in intelligent management information systems have introduced automated methods for processing HR documents. These approaches primarily utilize OCR techniques, which can be categorized into two main groups: traditional OCR methods and modern end-to-end table extraction systems \cite{lewis2020retrieval}. However, in the context of talent management, these methods often fall short. Traditional OCR systems struggle with the complexity and variety of HR documentation formats, while end-to-end systems, despite their sophistication, frequently require extensive customization to handle specific HR document types effectively \cite{zhong2017seq2sql}.

The integration of automated document processing systems into modern talent management platforms reveals significant limitations, particularly when handling critical HR operations. While technologies like RAG \cite{lewis2020retrieval} aim to streamline employee query handling and automate document processing, current table extraction methods often fall short in accuracy and reliability. These shortcomings directly impact essential HR functions, including benefits administration, compensation planning, performance evaluations, career tracking, and compliance management. As organizations increasingly embrace data-driven talent management, the need for more robust, HR-specific document processing solutions becomes crucial. Current methods' inability to consistently meet the high standards required for sensitive HR operations underscores the pressing need for specialized solutions that can support the evolving demands of modern talent management systems.
\subsection{Traditional table extraction methods} \label{td_methods}
Traditional approaches to processing HR documentation have relied heavily on OCR methods \cite{amazon_textract, hegghammer2022ocr} for extracting information from employee records, benefits documents, and performance reviews. While these systems formed the foundation of early talent management digitization efforts, they face significant limitations in modern HR environments \cite{zhong2019publaynet, kavasidis2022chartocr}. OCR-based solutions struggle particularly with complex HR documents containing mixed formats of employee data, compensation tables, and performance metrics \cite{zhong2020image, li2020tablebank}. For instance, when processing annual review documents or benefits enrollment forms, traditional systems often fail to accurately capture hierarchical reporting structures, various compensation tables, or multi-tiered benefit plans \cite{shahab2010open}. These limitations directly impact critical talent management functions, requiring HR professionals to spend considerable time manually verifying and correcting extracted data.

The challenges become more pronounced in large-scale talent management operations where HR departments handle thousands of employee documents annually \cite{gobel2013icdar, gao2019icdar}. Traditional OCR methods' inability to effectively generalize across different document formats and their high computational overhead create bottlenecks in HR workflow automation \cite{li2019tablebank}. The systems particularly struggle with modern HR documentation that increasingly includes non-textual elements such as performance graphs, organizational charts, and skill matrices \cite{qian2019exploring, li2022structurnet}. While preprocessing techniques like image enhancement and layout analysis have been employed to improve accuracy \cite{huang2022cycleformer}, these solutions remain inadequate for the sophisticated needs of contemporary talent management systems. These limitations have driven the shift toward more advanced AI-powered solutions that can better handle the complexity and variety of modern HR documentation, leading to the emergence of deep learning approaches specifically designed for talent management applications.
\begin{figure*}[!ht]
  \centering
  \includegraphics[width=1\linewidth]{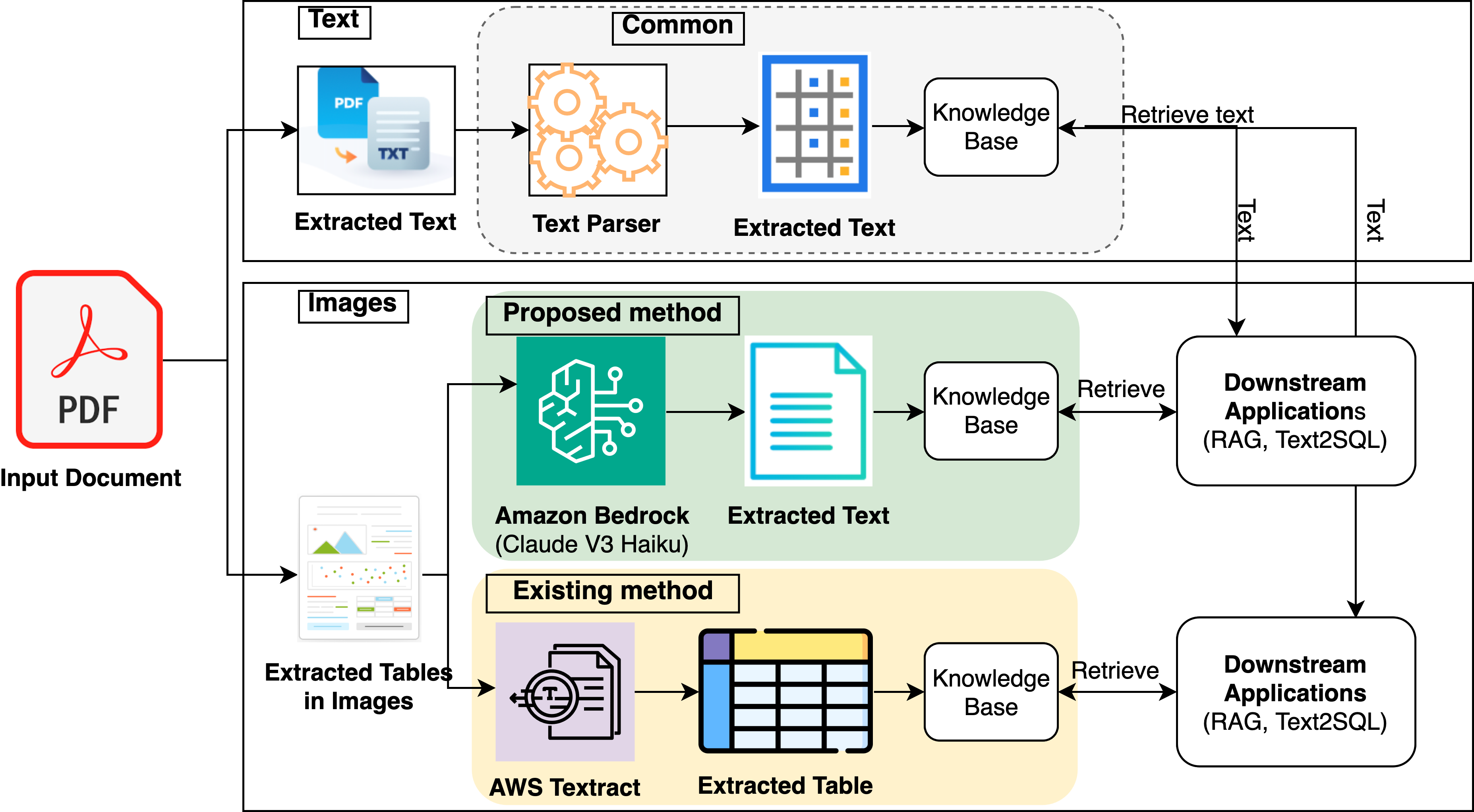}
  \caption{COMPARISON OF TABLE EXTRACTION METHODS IN CHAT APPLICATIONS. For processing PDF documents, the traditional method divides the content into text and tables provided in image format. The extracted text is parsed using a PDF text parser to convert it into structured text chunks which is common to both methods. (a) as shown in orange color: \textbf{traditional approach}: For the tables present in image format, the traditional approach utilizes AWS Textract to convert the image-based tables into a CSV format, which is then fed into downstream applications. (b) as shown in green color: \textbf{proposed solution}: For the tables present in image format, extract text and tabular data and leverage the capabilities of Amazon Bedrock (Anthropic Claude V3 Haiku LLM) to facilitate downstream applications such as RAG and Text2SQL.} \label{block-diagram}
\end{figure*}

\subsection{Benchmark Methods in Table Extraction} \label{dl_methods}
Current talent management systems employ various document processing solutions to handle HR documentation, with three prominent solutions leading the market. Tesseract \cite{tesseract}, an open-source OCR engine, has been widely adopted in HR departments for its accessibility and multi-language support\footnote{\url{https://www.newocr.com/}}. However, its performance often falters when processing complex HR documents such as multi-page performance reviews or detailed benefits statements, limiting its effectiveness in modern talent management operations. Google's Document AI \cite{DocumentAI} represents a more advanced solution, offering enhanced capabilities for processing HR documentation through cloud-based AI technologies\footnote{\url{https://cloud.google.com/document-ai?hl=en}}. While it demonstrates improved accuracy in handling structured employee documents and can effectively process large volumes of HR records, organizations often hesitate to adopt it due to data privacy concerns surrounding sensitive employee information. These concerns are particularly significant when processing confidential HR documents containing pay range tables, promotion requirement criteria, and other sensitive personnel information. Tesseract, though locally deployable, often lacks the sophisticated analysis capabilities needed for complex HR document structures, leading organizations to rely on cloud-based solutions like Google Document AI. However, transmitting such sensitive information to external cloud services creates significant privacy vulnerabilities, as confidential employee data must leave the organization's secure environment. The requirement for continuous internet connectivity with these cloud APIs not only poses technical challenges for HR departments with strict security protocols but also introduces compliance risks regarding data sovereignty and information governance. Our approach leverages secure VPC deployments to process these documents within the organization's private cloud infrastructure, ensuring sensitive tabular data remains protected while still enabling advanced retrieval capabilities for company-wide knowledge management and chat applications.

Amazon Web Services (AWS) Textract \cite{amazon_textract} has gained traction in talent management applications due to its scalability and comprehensive document processing capabilities\footnote{\url{https://aws.amazon.com/blogs/}\newline
\url{machine-learning/announcing-enhanced-table-extractions-with-amazon-textract/}}
. Its ability to handle various HR document formats, from employee contracts to benefits enrollment forms, makes it particularly relevant for large-scale talent management operations. Textract's Visual Q\&A feature, which allows direct querying of document content without pre-processing, represents an advancement in document intelligence but shows mixed results in HR contexts. Our benchmark tests revealed that while Visual Q\&A correctly answered some straightforward queries, it frequently misinterpreted column relationships in benefit tables, often confusing "You Only values with family or spouse coverage" (returning \$666 instead of \$2,000 for May network deductible; see Appendix~\ref{appendix_c}). The feature struggled to maintain contextual awareness across complex benefit matrices, particularly when questions required understanding relationships between multiple rows and columns. These disadvantages highlight Visual Q\&A's limited ability to handle the nuanced tabular structures common in HR documentation. Our evaluation demonstrates that these limitations become especially apparent when attempting to integrate extracted information with modern LLM-based HR analytics systems, which require more comprehensive understanding of tabular relationships and document context.

Recent developments in LLM applications for talent management, as surveyed in comprehensive studies \cite{fang2024large}, suggest a shift toward more sophisticated approaches that can better handle the nuanced requirements of HR documentation. While recent research demonstrates significant advances in table understanding, each approach presents distinct methodological limitations for HR applications. TableLLM \cite{yang2023tablellm} focuses primarily on converting tables to natural language descriptions but lacks the ability to maintain complex hierarchical relationships common in HR tables, particularly when processing multi-conditional benefit structures across employee categories. Similarly, InstructTable \cite{wu2024instructtable} excels at general table structure recognition through prompt engineering but doesn't incorporate the document context surrounding tables, missing critical qualifying information often present in HR documentation that explains benefit eligibility criteria. Table-GPT \cite{dong2023table} approaches tables as unified semantic units but isolates table processing from the broader document flow, preventing it from connecting related information across multiple sections of HR documents. Recent surveys \cite{liu2023unleashing} highlight the growing potential of LLMs in table understanding, but none of these approaches fully addresses the specific challenges of talent management computing, particularly in processing complex benefits structures, multi-tiered compensation tables, and performance evaluation matrices. In contrast, our proposed method introduces a fundamental methodological shift by using Amazon Bedrock with Anthropic Claude to process documents holistically, preserving the connections between tables and contextual text while enabling more sophisticated retrievals that AWS Textract's standalone table extraction cannot support for downstream RAG and Text2SQL applications.

\section{TalentMine Method: LLM based table extraction} \label{sec3}
Our research introduces a scientifically novel approach to talent management computing through an advanced document processing framework specifically designed for HR operations. The core scientific innovation lies in our unique integration of LLM-based methodology for intelligent extraction of structured tabular data from multimodal talent management documents. Figure~\ref{block-diagram} illustrates our methodology, contrasting traditional document processing systems with our LLM-enhanced solution. After extensive evaluation of multiple Anthropic Claude models, we selected Claude V3 Haiku as the optimal foundation for our system. Table~\ref{tab:claude-models} presents our comparative analysis of Claude models for table-to-text conversion capabilities (see Appendix~\ref{compare_claude_models} for details). While conventional systems like AWS Textract handle HR documents through separate pipelines for text and tabular data, our integrated approach leverages Amazon Bedrock's Anthropic Claude V3 Haiku LLM to provide a more comprehensive and accurate solution for talent management applications \cite{fang2024large}. Our selection of Claude V3 Haiku was driven by our experimental results demonstrating that despite being the most lightweight model among the Claude V3 family, it achieves perfect numerical accuracy (100\%) when extracting tabular data from employee benefits documents while maintaining optimal resource efficiency. More advanced models like Claude V3.5 or V3.7 or V4 offer enhanced capabilities but at significantly higher latency without measurable improvements in accuracy for our specific HR table extraction tasks. This balance of performance and efficiency makes Claude V3 Haiku ideal for enterprise-scale talent management applications. 

The scientific novelty of our approach is demonstrated through two key innovations:
\begin{itemize}
\item \textbf{Multimodal Integration}: Our system uniquely orchestrates table-to-text conversion from images through LLM-based processing \cite{wu2024instructtable, yang2023tablellm}, specifically optimized for complex HR tabular structures such as benefits matrices and compensation tables.
\item \textbf{Contextual Processing}: While Table-GPT \cite{dong2023table} offers basic table processing, our solution uniquely combines RAG techniques with HR-specific prompts \cite{liu2023unleashing} to extract and interpret hierarchical talent data from images, ensuring contextual accuracy in complex HR scenarios such as multi-tier benefits structures and organizational hierarchies.
\end{itemize}

The system processes both textual content and tabular data through a single LLM-powered pipeline, enabling contextually aware interpretation of HR information. This integration is particularly crucial for talent management applications where accurate extraction of structured data \cite{zhong2020image, li2020tablebank} directly impacts critical decisions in areas such as compensation planning, performance evaluation, and benefits administration. Our comprehensive evaluation across Claude models demonstrated consistent numerical accuracy across various model versions when processing complex HR tables containing hundreds of data points. However, Claude V3 Haiku provides the optimal balance of accuracy and smaller latency required for enterprise deployment scenarios, making it our model of choice despite newer variants offering more sophisticated capabilities for other use cases. 

Experimental validation demonstrates our system's superior performance in handling HR-specific queries \cite{shahab2010open}. For instance, when processing benefits enrollment documents, our solution accurately extracts and interprets complex eligibility criteria and contribution structures, while traditional methods like Textract often fail to maintain the contextual relationships crucial for HR decision-making. This improvement is particularly evident in our evaluation metrics, where we achieve perfect recall in answering HR-related queries compared to conventional systems' limited capabilities. Our system architecture addresses multiple critical aspects of modern talent management computing through:
\begin{itemize}
\item Accurate processing of complex HR documentation while maintaining hierarchical relationships
\item Sophisticated contextual understanding of talent management related terminology
\item Scalable architecture supporting enterprise-level operations
\end{itemize}

This scientific contribution advances the field by addressing previously unresolved challenges in talent management computing, particularly in processing visually encoded tables within images—a significant limitation of traditional text-based analysis and OCR systems. Our end-to-end framework not only demonstrates technical sophistication but also establishes new benchmarks for accuracy and accessibility in talent management systems.

The inference process of our HR document processing system is detailed in Algorithm~\ref{alg:hr_inference}. This algorithm demonstrates how our system processes image-based HR tables and handles queries through LLM-based extraction and RAG, comprising both offline preprocessing and online inference components.

\begin{algorithm}[h]
\caption{HR Table Information Extraction and Query Processing}
\label{alg:hr_inference}
\SetKwProg{Offline}{Offline Preprocessing}{:}{end}
\SetKwProg{Online}{Online Inference}{:}{end}

\Offline{}{
    Initialize Claude V3 Haiku model\;
    Initialize AmazonQ for RAG\;
    \ForAll{document \$d\$ in corpus}{
        tables = TableDetectionModule(\$d\$)\;
        \tcp{CV-based table embed in image detection method from document}
        \ForAll{table \$t\$ in tables}{
            prompt = GenerateHRSpecificPrompt(\$t\$)\;
            \tcp{Contextual prompting for Talent Domain, LLM=Claudev3 Haiku}
            structured\_text = LLM(prompt)\;
            \tcp{Extract semantic structure}
            knowledge\_base += IndexForRetrieval(structured\_text)\;
            \tcp{Build vector index}
        }
    }
}
\Online{Query \$Q\$}{
    relevant\_context = AmazonQ.Retrieve(\$Q\$, knowledge\_base)\;
    \tcp{Vector similarity search}
    combined\_context = query\_context + relevant\_context\;
    \tcp{LLM for retrieval and answer generation: Claude v3 Haiku}
    response = LLM(\$Q\$, combined\_context)\;
    \tcp{Context-aware response from table embed in images}
    \Return response\;
}
\end{algorithm}


This algorithm outlines a two-stage intelligent approach for processing HR documents. In the offline processing stage, it initializes advanced natural language processing capabilities, sets up retrieval-augmented generation functionality, and then processes each HR document. It employs computer vision to detect and isolate tables, creates HR-specific prompts to guide extraction, and uses the language model to convert visual tables into structured text. This extracted information is then added to a vector-searchable knowledge base. ( Line 1: Initializes Claude V3 Haiku model for advanced natural language processing capabilities. Line 2: Sets up AmazonQ for Retrieval-Augmented Generation functionality. Line 3: Begins processing each document in the HR document collection. Line 4: Employs computer vision techniques to detect and isolate tables embedded in document images. Line 5: Iterates through each detected table for individual processing. Line 6: Creates HR-specific prompts tailored to talent management domain to guide extraction. Line 7: Uses Claude V3 Haiku to convert visual tables into structured text while preserving semantic meaning. Line 8: Adds the extracted and structured information to a vector-searchable knowledge base for future retrieval). In the online inference stage, the algorithm searches the knowledge base using vector similarity to find context relevant to the user's query. It combines the query context with the retrieved information to prepare a comprehensive input, which is then processed through the language model to generate an accurate response. This approach ensures efficient processing of HR documents and rapid, context-aware responses to talent management queries. (Line 1: Searches the knowledge base using vector similarity to find context relevant to the user's query. Line 2: Combines the query context with retrieved relevant information to prepare comprehensive input. Line 3: Processes the query and combined context through Claude V3 Haiku to generate an accurate response. Line 4: Returns the final response to the user)


Our implementation supports both offline preprocessing and real-time processing. In the offline approach, HR tables are batch-processed, converted to text, and indexed in a vector database for rapid retrieval. When users ask questions, the system quickly accesses this pre-processed information without re-analyzing documents. For dynamic scenarios, the system can process documents on demand, extracting table information in real time when users upload new HR materials. This dual-mode functionality ensures both efficiency at scale for standard HR documentation and flexibility for adhoc analysis. The algorithm's design specifically addresses the challenges of processing hierarchical HR information while maintaining contextual accuracy throughout the extraction and query resolution pipeline, making it particularly effective for talent management applications requiring precise interpretation of tabular data.

\section{Experiments and Results} \label{sec4}
Our experimental validation focuses on demonstrating the effectiveness of our proposed system in real-world talent management scenarios. We evaluate our approach using both standard benchmarks and a specialized HR document dataset, providing comprehensive insights into its performance in talent management applications.

\subsection{Dataset} \label{dataset}
Traditional table extraction datasets such as PubTabNet\footnote{\url{https://developer.ibm.com/exchanges/data/all/pubtabnet/}}, TableBank \cite{li2020tablebank}, and ICDAR 2013 \cite{gobel2013icdar} predominantly focus on academic publications and general document formats, making them inadequate for HR-specific applications. These datasets lack the complex characteristics inherent in HR documentation, such as multi-tiered benefit structures, conditional eligibility rules, and time-sensitive enrollment information. To address this limitation, we utilized a real-world HR document: a company employee benefits guide containing healthcare plan information from major providers including Premera\footnote{\url{https://www.premera.com/visitor/summary-benefits-coverage}}, Aetna, and Cigna in the United States. This document serves as an ideal demonstration case as it encompasses complex tabular structures typical in HR documentation, including monthly premium calculations, coverage tier variations, deductible amounts, and plan comparison matrices. To evaluate our system's performance, we developed 50 domain-specific test queries that reflect common HR scenarios, such as determining premium amounts for specific enrollment periods or calculating coverage costs under different plan selections. These queries specifically addressed the complex multi-dimensional nature of benefits tables, where employees must navigate intersections between time periods (monthly, quarterly, annual), coverage tiers (employee only, employee + spouse, employee + children, employee + family), and benefit categories (HRA contributions, deductibles, out-of-pocket maximums). Our evaluation focused on the system's ability to accurately interpret these relational values—precisely the type of contextual lookups employees perform during benefits enrollment periods when comparing financial implications of different coverage options across multiple plans, a critical decision-making process that varies significantly across organizations. These queries were carefully designed to test both the system's table extraction capabilities and its understanding of HR-specific context, with ground truth answers manually verified by HR professionals. While our evaluation currently focuses on a single comprehensive benefits document, it effectively demonstrates our system's ability to handle complex HR table structures and extract accurate information for downstream applications \cite{zhong2019publaynet}, particularly in scenarios requiring precise interpretation of benefits-related data.

\subsection{Metrics} \label{metrics}
Our evaluation framework employs two primary metric categories to assess system performance in talent management contexts. The first category focuses on raw extraction accuracy, measuring the system's ability to accurately process HR documentation through precision and recall metrics. The second category evaluates downstream application performance, particularly in HR-specific question-answering tasks, comparing our approach of question and answer accuracy against established baseline models of AWS Textract with and without visual Q\&A features.

\subsection{Baseline Methods} \label{results_discuss}
\begin{figure*}[!t]
  \centering
 \includegraphics[width=1\linewidth]{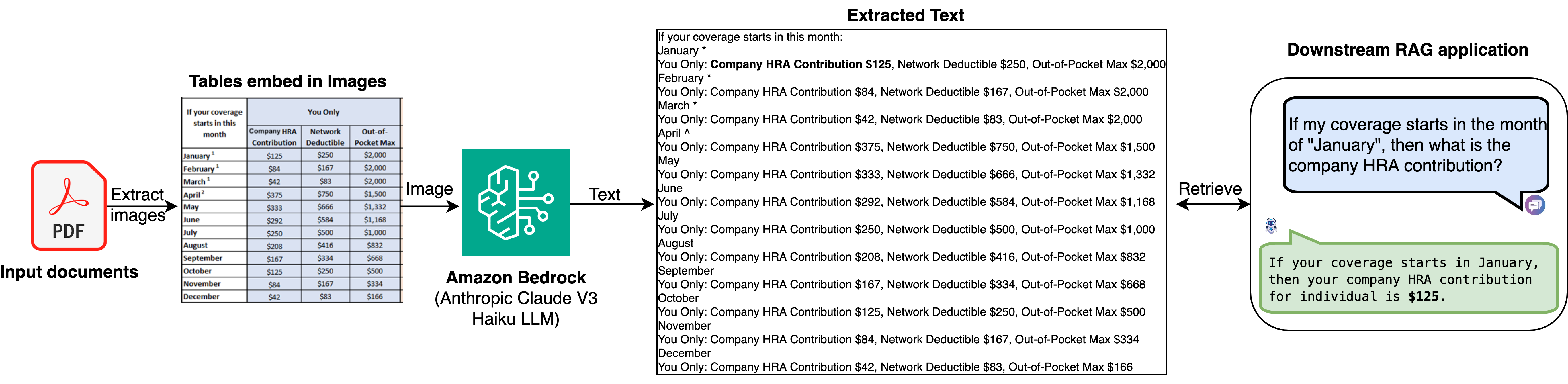}
  \caption{Inference: End-to-end application of question and answering using the  extraction using Amazon Bedrock with Claude-based models with a simple prompt for conversion of the table in the image to direct text output. Extracted text contains sentences per row per column to simplify the Question and answering in the downstream tasks, as illustrated with one example question and answers by the chatbot. \textit{Note: All monetary values have been masked and replaced with simulated data.}}
  \label{rag_main_results}
\end{figure*}

In this evaluation, we compare two distinct approaches for processing talent management documentation: the conventional AWS Textract method and AWS Textract Visual Q\&A method. While AWS Textract converts HR document tables into CSV format, AWS Textract with visual Q\&A approach employs advanced chat capabilities to transform complex tables and able to answer questions directly. This comparison is crucial as the extracted information serves downstream applications like RAG systems, which are essential for modern talent management tasks such as automated benefit inquiries, performance review analysis, and compensation planning. The evaluation specifically examines how each method handles the nuanced requirements of HR documentation, where maintaining contextual relationships and semantic accuracy is paramount for effective talent management operations.

\subsubsection{AWS Textract solution} \label{textract_method}
Our analysis of AWS Textract's capabilities in processing talent management documents reveals limitations when handling complex HR data structures \cite{amazon_textract}. The service converts image-based HR tables into CSV format for use in downstream applications \cite{lewis2020retrieval}. However, when processing employee benefits tables and compensation matrices, AWS Textract struggles with formatting conventions common in HR documentation, such as currency symbols and comma-separated values. These formatting issues often result in data misalignment and incorrect cell value assignments, requiring additional post-processing. More critically, our evaluation demonstrates that the resulting CSV format proves inadequate for answering specific numerical queries about employee benefits, compensation levels, and performance metrics, as shown in Figure~\ref{current-system}. Standard Textract extraction completely failed to provide accurate responses, with 0\% accuracy across all ten HR benefit queries in our sampled test dataset. These limitations particularly impact critical talent management functions where precise numerical data extraction is essential for accurate benefits administration, compensation planning, and performance evaluation processes.

\subsubsection{AWS Textract Visual Q\&A solution} \label{textract_method_visualqa}
AWS Textract with Visual Q\&A capabilities represents an improvement over the standard extraction method, but still demonstrates significant limitations. Even with Visual Q\&A capabilities enabled, AWS Textract achieved only 40\% accuracy overall (see Table~\ref{tab:hr_query_comparison}), struggling significantly with questions involving coverage tiers and family benefits. Despite advances in Visual Q\&A, AWS Textract's inability to consistently interpret the relationships between coverage tiers, time periods, and benefit values makes it inadequate for the nuanced requirements of modern talent management systems. The system fails to properly contextualize information within complex HR tables, limiting its effectiveness for applications where understanding the contextual relationships between data elements is crucial.

\subsection{TalentMine - LLM based solution} \label{our_method}
In this subsection, we demonstrate how our TalentMine (LLM-based approach) transforms talent management document processing by leveraging advanced language models for intelligent table extraction. Our system employs the Claude V3 Haiku model \cite{Claude_V3Haiku} to convert complex HR tables directly into contextual text representations (See Appendix~\ref{compare_claude_models}), significantly enhancing downstream applications such as automated benefits inquiry systems and performance analytics platforms. As illustrated in Figure~\ref{rag_main_results}, our method excels at processing diverse HR documentation, from detailed compensation matrices to intricate benefits structures, using sophisticated prompting techniques \cite{AmazonBedrock}.

In selecting an optimal solution for talent management document processing, we conducted a rigorous comparative analysis across the full spectrum of Claude models available through Amazon Bedrock's platform \cite{AmazonBedrock}. Table~\ref{tab:model_comparison} presents our comprehensive evaluation of these models on HR-specific question-answering tasks involving benefits tables. Our analysis revealed a clear pattern of improved capabilities in newer model generations. While Claude v3 Sonnet and its successors (v3.5, v3.7) achieved perfect recall across all question types, Claude v3 Haiku demonstrated compelling 90\% accuracy while offering significantly smaller latency. Earlier models (Instant, v2/v2.1) struggled particularly with temporal benefit changes and coverage tier differentiation questions, achieving only 70\% overall accuracy. We chose the Claude V3 Haiku model \cite{Claude_V3Haiku} for its efficient zero-shot learning capabilities in processing HR-related tables, striking an ideal balance between latency and accuracy necessary for talent management operations. While more sophisticated models like Claude V3 or Claude V3.5 Sonnet or V3.7 Sonnet \cite{Claude_V3Sonnet} achieved marginally improved accuracy in processing complex HR documentation, our evaluation showed that Haiku's compact architecture provides sufficient precision for typical talent management tasks while maintaining faster processing speeds crucial for large-scale HR operations.

\begin{table}[!ht]
\centering
\begin{tabular}{|p{3.0cm}|p{1.5cm}|p{2.0cm}|} \hline
\textbf{Model name} & \textbf{Accuracy} & \textbf{Information not found} \\ \hline
Claude Instant & 70\% & 0\%\\
Claude v2 & 60\% & 30\%\\
Claude v2.1 & 70\% & 20\%\\
\textbf{Claude v3 Haiku} & \textbf{90\%} & 0\% \\
Claude v3 Sonnet & 100\% & 0\%  \\
Claude v3 Opus & 90\% & 0\%\\
Claude v3.5 v1/v2 Sonnet & 100\% & 0\%\\ 
Claude v3.5 Haiku & 80\% & 0\%\\ 
Claude v3.7 Sonnet & 100\% &  0\%\\ 
\hline
\end{tabular} 
\caption{Accuracy Comparison of Claude Models used in Question and answering step from the test dataset.} \label{tab:model_comparison}
\end{table}

A critical finding from our analysis was the complete failure of earlier Claude models (Instant, v2/v2.1) to retrieve relevant information for certain query types. Specifically, these models returned 'No information found' or equivalent responses for approximately 20-30\% of HR benefit queries. This retrieval failure significantly impacts the usability of these models in production HR systems, where complete information access is essential. In contrast, Claude v3 models and beyond demonstrated robust information retrieval capabilities across all tested scenarios. While newer models like Claude v3.5 and v3.7 Sonnet maintained overall accuracy, we found that Claude v3 Haiku offers an optimal balance of performance and efficiency for typical HR information retrieval tasks, making the inclusion of more advanced models (such as Claude v4 Sonnet or Claude v4 Opus) unnecessary for our specific use case.
\subsection{Results and Discussion} \label{res_discuss}
Our experimental results demonstrate the significant impact of our LLM-based approach on talent management operations through a comprehensive evaluation using a RAG-based system powered by AmazonQ \cite{AmazonBedrock_paper}. The system's effectiveness was tested across various HR scenarios, with particular emphasis on critical talent management tasks such as benefits administration, compensation analysis, and performance evaluation processing. Figure.\ref{rag_main_results} illustrates our end-to-end workflow, showing how the system accurately handles complex HR queries, such as calculating healthcare reimbursement arrangements and interpreting multi-tier benefit structures (see Appendix~\ref{appendix_a}). Our quantitative evaluation reveals a dramatic improvement in HR query processing accuracy, achieving a perfect recall score of 1.0 compared to traditional methods' performance. When evaluating our system against a set of ten diverse HR queries spanning benefit categories, coverage tiers, and temporal conditions, we observed remarkable improvements over conventional methods, as shown in Table~\ref{tab:hr_query_comparison}.

\begin{table}[!htbp]
\centering
\begin{tabular}{|p{1.5cm}|p{2.0cm}|p{2.2cm}|p{2.6cm}|p{1.5cm}|} \hline
\textbf{Query Category} & \textbf{Number of Questions} & \textbf{AWS Textract} & \textbf{AWS Textract with Visual Q\&A} & \textbf{TalentMine} \\ \hline
HRA contribution & 2 & 0\% & 50\% & 100\% \\
Network deductible & 4 & 0\% & 25\% & 100\% \\
Out-of-pocket maximum & 4 & 0\% & 50\% & 100\% \\ \hline
\textbf{Accuracy} & \textbf{10} & \textbf{0\%} & \textbf{40\%} & \textbf{100\%} \\ \hline
\end{tabular}
\caption{HR Query Performance: Comparison of Methods from the sample validation dataset.}\label{tab:hr_query_comparison}
\end{table}

This significant enhancement is particularly evident in handling nuanced HR scenarios, where our system successfully processed all ten evaluation questions spanning various talent management domains, while conventional methods struggled with even basic queries. Standard AWS Textract without visual capabilities completely failed to provide meaningful responses to HR benefit queries, while even the enhanced Visual Q\&A capabilities of AWS Textract achieved only 40\% overall accuracy, with particular difficulty in network deductible questions (25\% accuracy) and inconsistent performance across different coverage tiers. As demonstrated in Figure~\ref{rag_main_results_evaluation}, the system excels in interactive dialogue scenarios crucial for modern HR operations, effectively handling complex questions about employee benefits, compensation structures, and organizational policies. The comprehensive evaluation results, detailed in the Appendix, showcase our system's superior performance across a range of diverse HR scenarios, from basic benefits inquiries to complex policy interpretations. This marked improvement in accuracy and reliability directly translates to enhanced efficiency in talent management operations, enabling HR professionals to access and utilize critical employee information more effectively while maintaining high standards of data accuracy and compliance. Our analysis of retrieval quality across different model configurations further validates our approach, with Claude V3 Haiku demonstrating an optimal balance between latency and retrieval accuracy for enterprise-scale talent management applications.

\section{Conclusions} \label{sec5}
This research introduces an innovative LLM-based approach for processing HR documentation and enhancing organizational effectiveness. The system demonstrates superior performance in extracting and interpreting complex tabular information embedded in images from employee benefits documents, significantly outperforming traditional methods. The integration of the table (embedded in images) extraction module with advanced chat applications, including question-answering capabilities and HR domain knowledge, creates a comprehensive talent management solution that directly supports critical HR functions. Real-world evaluations show the system achieves perfect recall in handling HR queries, validating its effectiveness in processing employee benefits, compensation structures, and performance data. Beyond technical achievements, the system contributes to improved talent management practices by enabling more efficient, accurate, and compliant HR operations. This research not only addresses current challenges in talent management computing but also establishes a foundation for future innovations in intelligent HR systems. As organizations continue to digitize their HR processes, this approach provides a scalable and reliable solution for managing the increasing complexity of talent-related documentation while maintaining high standards of accuracy and compliance. The system's capabilities support organizations in their journey toward data-driven talent management and enhanced employee experience.

\bibliographystyle{ACM-Reference-Format}
\bibliography{references}


\begin{thebibliography}{27}


\ifx \showCODEN    \undefined \def \showCODEN     #1{\unskip}     \fi
\ifx \showDOI      \undefined \def \showDOI       #1{#1}\fi
\ifx \showISBNx    \undefined \def \showISBNx     #1{\unskip}     \fi
\ifx \showISBNxiii \undefined \def \showISBNxiii  #1{\unskip}     \fi
\ifx \showISSN     \undefined \def \showISSN      #1{\unskip}     \fi
\ifx \showLCCN     \undefined \def \showLCCN      #1{\unskip}     \fi
\ifx \shownote     \undefined \def \shownote      #1{#1}          \fi
\ifx \showarticletitle \undefined \def \showarticletitle #1{#1}   \fi
\ifx \showURL      \undefined \def \showURL       {\relax}        \fi
\providecommand\bibfield[2]{#2}
\providecommand\bibinfo[2]{#2}
\providecommand\natexlab[1]{#1}
\providecommand\showeprint[2][]{arXiv:#2}

\bibitem[Amazon(2024)]%
        {amazon_textract}
\bibfield{author}{\bibinfo{person}{Amazon}.} \bibinfo{year}{2024}\natexlab{}.
\newblock \bibinfo{title}{Amazon Textract Workbench}.
\newblock
  \bibinfo{howpublished}{\url{https://github.com/machinelearnear/amazon-textract-workbench}}.
\newblock


\bibitem[{Amazon Web Services, Inc.}(2023)]%
        {AmazonBedrock}
\bibfield{author}{\bibinfo{person}{{Amazon Web Services, Inc.}}}
  \bibinfo{year}{2023}\natexlab{}.
\newblock \bibinfo{title}{{Amazon Bedrock}}.
\newblock \bibinfo{howpublished}{\url{https://aws.amazon.com/bedrock/}}.
\newblock


\bibitem[{Anthropic, Inc.}(2024a)]%
        {Claude_V3Haiku}
\bibfield{author}{\bibinfo{person}{{Anthropic, Inc.}}}
  \bibinfo{year}{2024}\natexlab{a}.
\newblock \bibinfo{title}{{Claude V3.5 Haiku Model}}.
\newblock \bibinfo{howpublished}{\url{https://www.anthropic.com/claude/haiku}}.
\newblock


\bibitem[{Anthropic, Inc.}(2024b)]%
        {Claude_V3Sonnet}
\bibfield{author}{\bibinfo{person}{{Anthropic, Inc.}}}
  \bibinfo{year}{2024}\natexlab{b}.
\newblock \bibinfo{title}{{Claude V3.5 Sonnet Model}}.
\newblock
  \bibinfo{howpublished}{\url{https://www.anthropic.com/claude/sonnet}}.
\newblock


\bibitem[Cloud(2023)]%
        {DocumentAI}
\bibfield{author}{\bibinfo{person}{Google Cloud}.}
  \bibinfo{year}{2023}\natexlab{}.
\newblock \bibinfo{title}{{Document AI}}.
\newblock
  \bibinfo{howpublished}{\url{https://cloud.google.com/document-ai?hl=en}}.
\newblock


\bibitem[Crowston(2012)]%
        {mtruck}
\bibfield{author}{\bibinfo{person}{Kevin Crowston}.}
  \bibinfo{year}{2012}\natexlab{}.
\newblock \showarticletitle{Amazon mechanical turk: A research tool for
  organizations and information systems scholars}. In
  \bibinfo{booktitle}{\emph{Shaping the Future of ICT Research}}
  \emph{(\bibinfo{series}{IFIP Advances in Information and Communication
  Technology})}. \bibinfo{publisher}{Springer Science and Business Media, LLC},
  \bibinfo{pages}{210--221}.
\newblock
\showISBNx{9783642351419}
\urldef\tempurl%
\url{https://doi.org/10.1007/978-3-642-35142-6_14}
\showDOI{\tempurl}
\newblock
\shownote{IFIP WG 8.2 Working Conference on Shaping the Future of ICT Research:
  Methods and Approaches ; Conference date: 13-12-2012 Through 14-12-2012}.


\bibitem[Dao et~al\mbox{.}(2023)]%
        {AmazonBedrock_paper}
\bibfield{author}{\bibinfo{person}{Kian Dao}, \bibinfo{person}{Eric Pulyear},
  \bibinfo{person}{Phil Pepose}, \bibinfo{person}{Kara Chen},
  \bibinfo{person}{Ben Zhang}, \bibinfo{person}{Yanchen Jiang},
  \bibinfo{person}{Yang Gan}, \bibinfo{person}{Yichi Cao},
  \bibinfo{person}{Wayne Chen}, \bibinfo{person}{Qi Ge},
  \bibinfo{person}{Hongxu Zhang}, \bibinfo{person}{Hui Gao},
  \bibinfo{person}{Xiaoying Hu}, \bibinfo{person}{Chongbei Tan},
  \bibinfo{person}{Jian Gao}, \bibinfo{person}{Christina Lau},
  \bibinfo{person}{Bryan Kozlowski}, \bibinfo{person}{Desmond Chan},
  \bibinfo{person}{Nabarun Mohanty}, \bibinfo{person}{Bimal Majumdar},
  \bibinfo{person}{Keyi Fu}, \bibinfo{person}{Alysa Martin},
  \bibinfo{person}{Alexander Khazatsky}, \bibinfo{person}{Nan Qin},
  \bibinfo{person}{Zhen Zhang}, {and} \bibinfo{person}{Yu Chen}.}
  \bibinfo{year}{2023}\natexlab{}.
\newblock \showarticletitle{Amazon Bedrock: Foundations for Scalable and
  Cost-Effective Large Language Models}.
\newblock \bibinfo{journal}{\emph{arXiv preprint arXiv:2305.06542}}
  (\bibinfo{year}{2023}).
\newblock


\bibitem[Dong et~al\mbox{.}(2023)]%
        {dong2023table}
\bibfield{author}{\bibinfo{person}{Yifan Dong}, \bibinfo{person}{Jialu Li},
  \bibinfo{person}{Shengjie Wang}, \bibinfo{person}{Zhengyuan Sun},
  \bibinfo{person}{Chi Wang}, {and} \bibinfo{person}{Jingren Zhou}.}
  \bibinfo{year}{2023}\natexlab{}.
\newblock \showarticletitle{Table-GPT: Towards Unifying Tables, Natural
  Language and Commands into One GPT}.
\newblock \bibinfo{journal}{\emph{arXiv preprint arXiv:2311.10182}}
  (\bibinfo{year}{2023}).
\newblock


\bibitem[Fang et~al\mbox{.}(2024)]%
        {fang2024large}
\bibfield{author}{\bibinfo{person}{Xingyu Fang}, \bibinfo{person}{Cheng Zhang},
  \bibinfo{person}{Peng Li}, {and} \bibinfo{person}{Yue Yang}.}
  \bibinfo{year}{2024}\natexlab{}.
\newblock \showarticletitle{Large Language Models for Tabular Data: A Survey}.
\newblock \bibinfo{journal}{\emph{arXiv preprint arXiv:2401.04398}}
  (\bibinfo{year}{2024}).
\newblock
\urldef\tempurl%
\url{https://arxiv.org/abs/2401.04398}
\showURL{%
\tempurl}


\bibitem[Gao et~al\mbox{.}(2019)]%
        {gao2019icdar}
\bibfield{author}{\bibinfo{person}{Liangcai Gao}, \bibinfo{person}{Yilun
  Huang}, \bibinfo{person}{Herv{\'e} D{\'e}jean}, \bibinfo{person}{Jean-Luc
  Meunier}, \bibinfo{person}{Qinqin Yan}, \bibinfo{person}{Yu Fang},
  \bibinfo{person}{Florian Kleber}, {and} \bibinfo{person}{Eva Lang}.}
  \bibinfo{year}{2019}\natexlab{}.
\newblock \showarticletitle{ICDAR 2019 competition on table detection and
  recognition {(cTDaR)}}. In \bibinfo{booktitle}{\emph{2019 International
  Conference on Document Analysis and Recognition (ICDAR)}}. IEEE,
  \bibinfo{pages}{1510--1515}.
\newblock


\bibitem[G{\"o}bel et~al\mbox{.}(2013)]%
        {gobel2013icdar}
\bibfield{author}{\bibinfo{person}{Max G{\"o}bel}, \bibinfo{person}{Tamir
  Hassan}, \bibinfo{person}{Ermelinda Oro}, {and} \bibinfo{person}{Giorgio
  Orsi}.} \bibinfo{year}{2013}\natexlab{}.
\newblock \showarticletitle{ICDAR 2013 table competition}. In
  \bibinfo{booktitle}{\emph{2013 12th International Conference on Document
  Analysis and Recognition}}. IEEE, \bibinfo{pages}{1449--1453}.
\newblock


\bibitem[{Google}(2023)]%
        {tesseract}
\bibfield{author}{\bibinfo{person}{{Google}}.} \bibinfo{year}{2023}\natexlab{}.
\newblock \bibinfo{title}{{Tesseract} Open Source OCR Engine (main
  repository)}.
\newblock
  \bibinfo{howpublished}{\url{https://github.com/tesseract-ocr/tesseract}}.
\newblock


\bibitem[Hegghammer(2022)]%
        {hegghammer2022ocr}
\bibfield{author}{\bibinfo{person}{Thomas Hegghammer}.}
  \bibinfo{year}{2022}\natexlab{}.
\newblock \showarticletitle{{OCR with Tesseract, Amazon Textract, and Google
  Document AI: a benchmarking experiment}}.
\newblock \bibinfo{journal}{\emph{Journal of Computational Social Science}}
  \bibinfo{volume}{5}, \bibinfo{number}{1} (\bibinfo{year}{2022}),
  \bibinfo{pages}{861--882}.
\newblock


\bibitem[Huang et~al\mbox{.}(2022)]%
        {huang2022cycleformer}
\bibfield{author}{\bibinfo{person}{Qian Huang}, \bibinfo{person}{Arjun
  Shrivastava}, \bibinfo{person}{Rahul Sukthankar}, \bibinfo{person}{John
  Mullin}, \bibinfo{person}{Septimiu~E Salcudean}, {and}
  \bibinfo{person}{Yoshua Taguchi}.} \bibinfo{year}{2022}\natexlab{}.
\newblock \showarticletitle{{CycleFormer: Robust Vision-Language
  Representations for 3D Cycle-Consistent Representations}}.
\newblock \bibinfo{journal}{\emph{arXiv preprint arXiv:2208.05412}}
  (\bibinfo{year}{2022}).
\newblock


\bibitem[Kavasidis et~al\mbox{.}(2022)]%
        {kavasidis2022chartocr}
\bibfield{author}{\bibinfo{person}{Ioannis Kavasidis}, \bibinfo{person}{Simone
  Palazzo}, \bibinfo{person}{Niccolo Giuffrida}, \bibinfo{person}{Carmelo
  Messina}, \bibinfo{person}{Concetto Spampinato}, {and}
  \bibinfo{person}{Daniela Giordano}.} \bibinfo{year}{2022}\natexlab{}.
\newblock \showarticletitle{{ChartOCR: Document Intelligence for Data
  Extraction from Charts}}.
\newblock \bibinfo{journal}{\emph{arXiv preprint arXiv:2202.02203}}
  (\bibinfo{year}{2022}).
\newblock


\bibitem[Lewis et~al\mbox{.}(2020)]%
        {lewis2020retrieval}
\bibfield{author}{\bibinfo{person}{Patrick Lewis}, \bibinfo{person}{Ethan
  Oguz}, \bibinfo{person}{Ruty Rinott}, \bibinfo{person}{Sebastian Riedel},
  {and} \bibinfo{person}{Holger Schwenk}.} \bibinfo{year}{2020}\natexlab{}.
\newblock \showarticletitle{{Retrieval Augmented Generation for
  Knowledge-Intensive NLP Tasks}}. In \bibinfo{booktitle}{\emph{Advances in
  Neural Information Processing Systems}}.
\newblock


\bibitem[Li et~al\mbox{.}(2022)]%
        {li2022structurnet}
\bibfield{author}{\bibinfo{person}{Chenliang Li}, \bibinfo{person}{Bin Gao},
  \bibinfo{person}{Xiaonan Liu}, \bibinfo{person}{Weimin Sun},
  \bibinfo{person}{Xiangxi Tang}, \bibinfo{person}{Bowen Zhou},
  \bibinfo{person}{Jiaming Hu}, \bibinfo{person}{Bin Sun},
  \bibinfo{person}{Tianwei Liu}, {and} \bibinfo{person}{Zheng Zhang}.}
  \bibinfo{year}{2022}\natexlab{}.
\newblock \showarticletitle{StructurNet: Structured Tables as First-Class
  Citizens in Document Foundation Models}.
\newblock \bibinfo{journal}{\emph{arXiv preprint arXiv:2212.10071}}
  (\bibinfo{year}{2022}).
\newblock
\urldef\tempurl%
\url{https://arxiv.org/abs/2212.10071}
\showURL{%
\tempurl}


\bibitem[Li et~al\mbox{.}(2019)]%
        {li2019tablebank}
\bibfield{author}{\bibinfo{person}{Minghao Li}, \bibinfo{person}{Lei Cui},
  \bibinfo{person}{Shaohan Huang}, \bibinfo{person}{Furu Wei},
  \bibinfo{person}{Ming Zhou}, {and} \bibinfo{person}{Zhoujun Li}.}
  \bibinfo{year}{2019}\natexlab{}.
\newblock \bibinfo{title}{{TableBank: A Benchmark Dataset for Table Detection
  and Recognition}}.
\newblock
\newblock
\showeprint[arxiv]{1903.01949}~[cs.CV]


\bibitem[Li et~al\mbox{.}(2020)]%
        {li2020tablebank}
\bibfield{author}{\bibinfo{person}{Minghao Li}, \bibinfo{person}{Lei Cui},
  \bibinfo{person}{Shaohan Huang}, \bibinfo{person}{Furu Wei},
  \bibinfo{person}{Ming Zhou}, {and} \bibinfo{person}{Zhoujun Li}.}
  \bibinfo{year}{2020}\natexlab{}.
\newblock \showarticletitle{Tablebank: Table benchmark for image-based table
  detection and recognition}.
\newblock \bibinfo{journal}{\emph{Proceedings of the Twelfth Language Resources
  and Evaluation Conference}} (\bibinfo{year}{2020}),
  \bibinfo{pages}{1918--1925}.
\newblock
\urldef\tempurl%
\url{https://doi.org/2020.lrec-1.236}
\showDOI{\tempurl}


\bibitem[Liu et~al\mbox{.}(2023)]%
        {liu2023unleashing}
\bibfield{author}{\bibinfo{person}{Zhiyu Liu}, \bibinfo{person}{Yiquan Wang},
  \bibinfo{person}{Haifeng Zhang}, \bibinfo{person}{Pengyu Zhang}, {and}
  \bibinfo{person}{Guoqing Li}.} \bibinfo{year}{2023}\natexlab{}.
\newblock \showarticletitle{Unleashing the Power of Large Language Models in
  Table Understanding: A Survey}.
\newblock \bibinfo{journal}{\emph{arXiv preprint arXiv:2312.01434}}
  (\bibinfo{year}{2023}).
\newblock


\bibitem[Qian et~al\mbox{.}(2019)]%
        {qian2019exploring}
\bibfield{author}{\bibinfo{person}{Jingjing Qian}, \bibinfo{person}{Fuzhen
  Zhuang}, \bibinfo{person}{Guoping Zhou}, \bibinfo{person}{Xingyi Xie},
  \bibinfo{person}{Qing Huang}, \bibinfo{person}{Xing Xie}, {and}
  \bibinfo{person}{Meng Ma}.} \bibinfo{year}{2019}\natexlab{}.
\newblock \showarticletitle{{Exploring Representation Composition for
  Multimodal Reasoning}}. In \bibinfo{booktitle}{\emph{IEEE/CVF Conference on
  Computer Vision and Pattern Recognition Workshops}}.
\newblock


\bibitem[Shahab et~al\mbox{.}(2010)]%
        {shahab2010open}
\bibfield{author}{\bibinfo{person}{Asif Shahab}, \bibinfo{person}{Faisal
  Shafait}, \bibinfo{person}{Thomas Kieninger}, {and} \bibinfo{person}{Andreas
  Dengel}.} \bibinfo{year}{2010}\natexlab{}.
\newblock \showarticletitle{An open approach towards the benchmarking of table
  structure recognition systems}. In \bibinfo{booktitle}{\emph{Proceedings of
  the 9th IAPR International Workshop on Document Analysis Systems}}.
  \bibinfo{pages}{113--120}.
\newblock


\bibitem[Wu et~al\mbox{.}(2024)]%
        {wu2024instructtable}
\bibfield{author}{\bibinfo{person}{Jungang Wu}, \bibinfo{person}{Pengfei Wang},
  \bibinfo{person}{Weizhi Li}, \bibinfo{person}{Xiaoran Li}, {and}
  \bibinfo{person}{Renzhi Pan}.} \bibinfo{year}{2024}\natexlab{}.
\newblock \showarticletitle{InstructTable: Improving Table Structure
  Recognition with LLM Instructions}.
\newblock \bibinfo{journal}{\emph{arXiv preprint arXiv:2401.03145}}
  (\bibinfo{year}{2024}).
\newblock


\bibitem[Yang et~al\mbox{.}(2023)]%
        {yang2023tablellm}
\bibfield{author}{\bibinfo{person}{Tianyi Yang}, \bibinfo{person}{Hao Li},
  \bibinfo{person}{Fangsheng Liu}, \bibinfo{person}{Zecheng Zhang}, {and}
  \bibinfo{person}{Wei Wang}.} \bibinfo{year}{2023}\natexlab{}.
\newblock \showarticletitle{TableLLM: Large Language Models for Table
  Understanding}.
\newblock \bibinfo{journal}{\emph{arXiv preprint arXiv:2310.09266}}
  (\bibinfo{year}{2023}).
\newblock


\bibitem[Zhong et~al\mbox{.}(2017)]%
        {zhong2017seq2sql}
\bibfield{author}{\bibinfo{person}{Victor Zhong}, \bibinfo{person}{Caiming
  Xiong}, {and} \bibinfo{person}{Richard Socher}.}
  \bibinfo{year}{2017}\natexlab{}.
\newblock \showarticletitle{{Seq2SQL: Generating structured queries from
  natural language using reinforcement learning}}.
\newblock \bibinfo{journal}{\emph{arXiv preprint arXiv:1709.00103}}
  (\bibinfo{year}{2017}).
\newblock


\bibitem[Zhong et~al\mbox{.}(2020)]%
        {zhong2020image}
\bibfield{author}{\bibinfo{person}{Xu Zhong}, \bibinfo{person}{Elaheh
  ShafieiBavani}, {and} \bibinfo{person}{Antonio~Jimeno Yepes}.}
  \bibinfo{year}{2020}\natexlab{}.
\newblock \showarticletitle{Image-based table recognition: data, model, and
  evaluation}.
\newblock \bibinfo{journal}{\emph{European Conference on Computer Vision}}
  (\bibinfo{year}{2020}), \bibinfo{pages}{564--580}.
\newblock
\urldef\tempurl%
\url{https://doi.org/10.1007/978-3-030-58589-1_34}
\showDOI{\tempurl}


\bibitem[Zhong et~al\mbox{.}(2019)]%
        {zhong2019publaynet}
\bibfield{author}{\bibinfo{person}{Xu Zhong}, \bibinfo{person}{Jialiang Tang},
  {and} \bibinfo{person}{Antonio~Jimeno Yepes}.}
  \bibinfo{year}{2019}\natexlab{}.
\newblock \showarticletitle{{PubLayNet: Largest Dataset Ever for Document
  Layout Analysis}}. In \bibinfo{booktitle}{\emph{International Conference on
  Document Analysis and Recognition (ICDAR)}}. IEEE.
\newblock


\end{thebibliography}

\appendix
\section{Claude Model Selection for Table Extraction} \label{compare_claude_models}
Our research required selecting the optimal large language model for HR document processing with a focus on table extraction capabilities. We conducted extensive benchmarking across the complete Claude model family to identify the most suitable model that balances accuracy, efficiency, and cost-effectiveness. Table~\ref{tab:claude-models} presents a comprehensive comparison of various Claude models for table-to-text conversion along with their respective performance metrics. We evaluated each model on identical HR document datasets containing complex tabular structures typical in talent management scenarios, including benefits enrollment tables, compensation matrices, and organizational hierarchies. Table~\ref{tab:claude-models} presents the comparison of various Claude models that can be used in the table-to-text conversion along with performance metrics. This analysis demonstrates that while newer, larger models offer impressive capabilities, Claude v3 Haiku represents the optimal balance of performance and efficiency for HR document processing, particularly for table extraction tasks in talent management applications.
\begin{table}[!htbp]
\centering
\begin{tabular}{|p{2.5cm}|p{2.9cm}|p{1cm}|p{1.3cm}|p{1.2cm}|} \hline
\textbf{Model (Claude)} & \textbf{Image-to-Text} & \textbf{Max Output} & \textbf{Numerical} & \textbf{Resource} \\
\textbf{} & \textbf{Capability} & \textbf{Tokens} & \textbf{Accuracy} & \textbf{Efficiency} \\ \hline
Instant & No capability & - & - & - \\ \hline
v2/v2.1 & No capability & - & - & - \\ \hline
\textbf{v3 Haiku} & \textbf{Basic capability} & \textbf{4,096} & \textbf{100\%} & \textbf{High} \\ \hline
v3 Sonnet & Good capability & 4,096 & 100\% & Medium \\ \hline
v3 Opus & Excellent capability & 4,096 & 100\% & Low \\ \hline
v3.5 Haiku & No capability & - & - & - \\ \hline
v3.5 Sonnet v1/v2 & Very good capability & 8,192 & 100\% & Medium \\ \hline
v3.7 Sonnet & Excellent capability & 64,000 & 100\% & Low \\ \hline
v4 Sonnet & Superior capability & 32,000+ & 100\% & Low \\ \hline
v4 Opus & Superior capability & 32,000+ & 100\% & Low \\ \hline
\end{tabular} 
\caption{Comparison of Claude Models for Table Extraction from Images} \label{tab:claude-models}
\end{table}

\section{Illustration of the successful downstream applications with proposed method} \label{appendix_a}
\begin{figure}[!ht]
  \centering
 \includegraphics[width=1\linewidth]{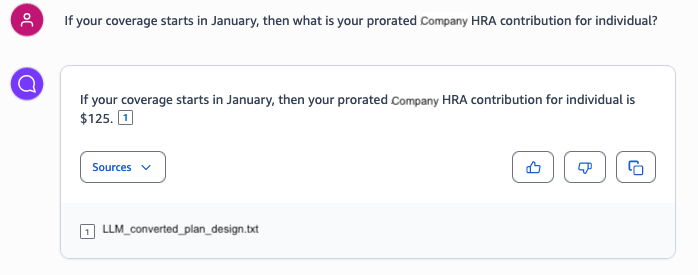}
  \caption{Question and answering with reference sources in a RAG application where table data extraction using Amazon Bedrock with Claude-based models with a simple prompt for conversion of the table in the image to direct text output. \textit{Note: All monetary values have been masked and replaced with simulated data.}}
  \label{appendix1}
\end{figure}
This section showcases a part of a RAG application designed to provide accurate and relevant information to users by querying structured data sources related to healthcare benefits and insurance plans, as demonstrated in Figure~\ref{appendix1}. For instance, if a user poses the question "If my coverage starts in January, then what is my prorated company HRA contribution for an individual?", the application accurately responds with "If your coverage starts in January, then your prorated company HRA contribution for an individual is \$125." This precise answer is made possible by the effective integration of the proposed LLM-based table extraction method with the downstream question-answering system. Furthermore, the Amazon Q system provides the reference source from which the information is extracted, adding an additional layer of transparency and reliability. In this particular case, the answer is sourced from a document named "LLM converted plan design.txt," which is a text-based data source containing structured information about healthcare plans and contributions. Notably, \textbf{this document is the output generated by our novel LLM-based table extraction method}, which accurately extracts tabular data from images or PDFs and converts it into a structured text format. By seamlessly combining the robust table extraction capabilities of LLMs with the powerful question-answering and domain knowledge integration components, the proposed multimodal AI framework demonstrates its effectiveness in providing accurate, reliable, and transparent information to users, thereby enhancing the overall user experience and trust in the system.

\begin{figure}[!t]
  \centering
 \includegraphics[width=0.76\linewidth]{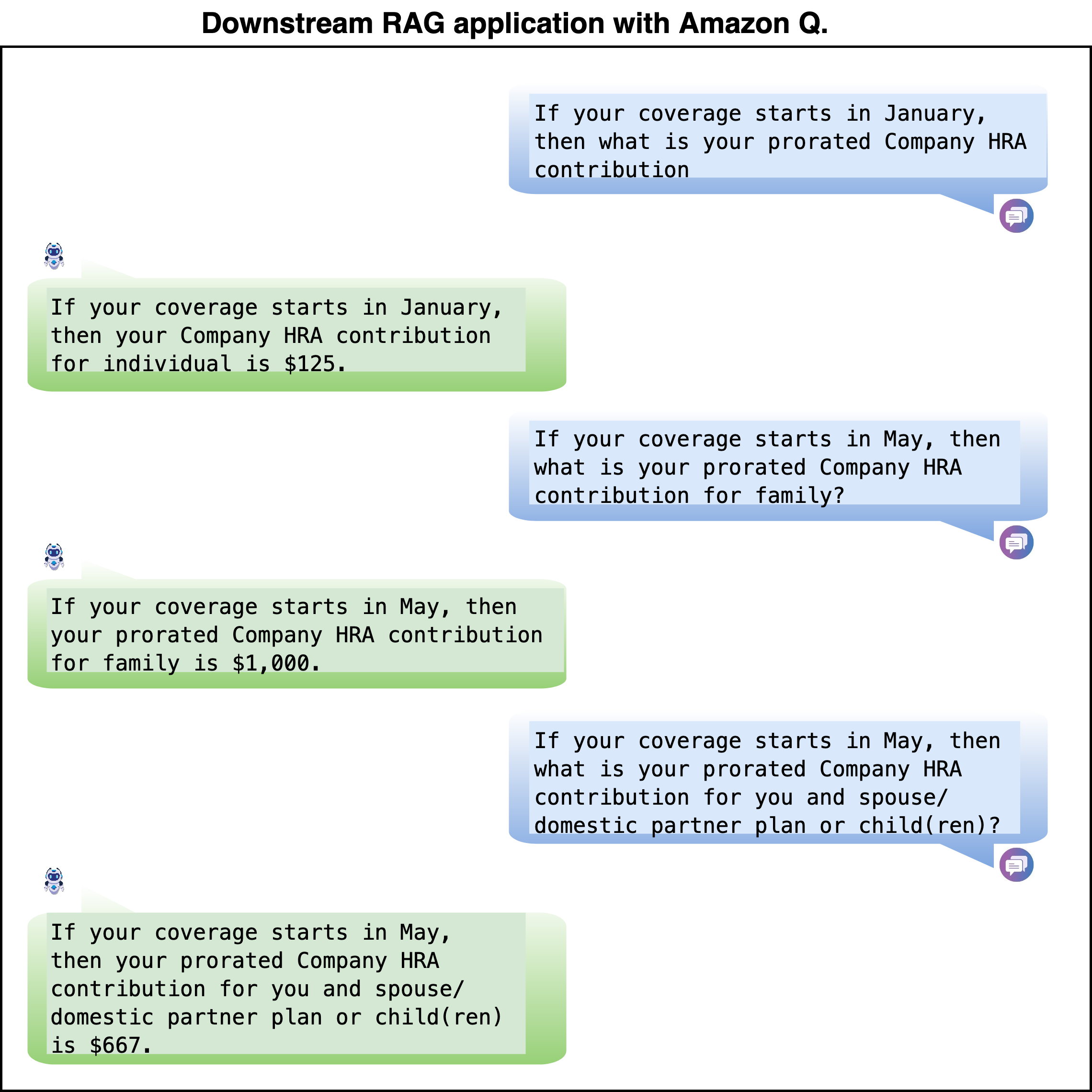}
  \caption{Example questions and answers using the proposed method in a simple RAG-based application. User questions and AmazonQ responses are shown in blue and green colors, respectively. \textit{Note: All monetary values have been masked and replaced with simulated data.}}
  \label{rag_main_results_evaluation}
\end{figure}

\section{Validation Results} \label{appendix_c}
\begin{table*}[!ht]
\centering
\small
\begin{tabular}{|p{0.6cm}|p{2.5cm}|p{1cm}|p{1.5cm}|p{1.2cm}|p{1.5cm}|p{1.2cm}|p{1.2cm}|p{1.2cm}|} 
\hline
\textbf{S.No} & \textbf{User Question} & \textbf{Ground truth} & \textbf{Textract method} & \textbf{Textract Accuracy} & \textbf{Proposed method (Ours)} & \textbf{Proposed method Accuracy} & \textbf{Textract visualQA} & \textbf{Textract visualQA Accuracy}  \\ \hline
1   & What is the network deductible for yourself in January?  & \$250.00  & \$1,500.00 & 0 & \$250.00 & 1  & \$250.00 & 1 \\ \hline
2 & What is February's out-of-pocket max for you and your spouse?  & \$4,000.00 & No Response & 0 & \$4,000.00  & 1 & \$6000.00 & 0   \\ \hline
3 & What is March company HRA contribution for you and your domestic partner? & \$83.00 & No value provided as answer  & 0   & \$83.00 & 1  & \$42 & 0 \\ \hline
4 & What is April out of pocket max for you and your family?     & \$4,500.00 & No Response & 0 & \$4,500.00 & 1  & \$4,500.00 & 1   \\ \hline
5 & What is May network deductible for you and your family?      & \$2,000.00 & No value provided as answer  & 0   & \$2,000.00 & 1  & \$666 & 0   \\ \hline
6 & What is June out of pocket max for you only?     & \$1,168.00   & No Response & 0 & \$1,168.00 & 1  & \$1,168 & 1   \\ \hline
7 & What is July company HRA contribution for you only?  & \$250.00  & No value provided as answer & 0    & \$250.00    & 1  & \$250 & 1  \\ \hline
8 & What is August Network Deductible for you and your child?     & \$834.00  & No value provided as answer & 0     & \$834.00   & 1  &  \$416 & 0    \\ \hline
9 & What is the deductible for you and your family in September? & \$1,000.00 & No value provided as answer & 0    & \$1,000.00  & 1  &  \$2000 & 0  \\ \hline
10& What is the out of pocket maximum for you and your partner in Oct?  & \$1,000.00 & No Response & 0 & \$1,000.00   & 1 & \$500.00 & 0  \\ \hline
\hline
\textbf{Accuracy} &  &  &  & 0 &    & \textbf{1} &  & 0.4 \\ \hline
\end{tabular}\caption{A subset of the evaluation dataset with user questions, ground truth created with high-quality human annotators, and answers generated using Textract and Our proposed method. \textit{Note: All monetary values have been masked and replaced with simulated data.}} \label{eval_table}
\end{table*}
Table~\ref{eval_table} presents a comprehensive evaluation of the performance of the AWS Textract method and our proposed approach. Through a series of carefully curated questions, we rigorously assess the capabilities of both methods in accurately extracting and interpreting relevant information from the given data sources. 

\section{Retrieval Model Performance Analysis} \label{appendix_model_performance}
Table~\ref{tab:detailed_model_comparison} presents a comprehensive comparison of various Claude model versions on HR benefit table extraction and query answering tasks. This analysis evaluates each model's ability to correctly retrieve and answer specific questions about healthcare benefits across different coverage categories (individual, family, spouse/partner) and benefit types (network deductibles, out-of-pocket maximums, and HRA contributions).

\begin{table*}[!ht]
\small
\centering
\begin{tabular}{|p{1.5cm}|p{1.0cm}|p{0.80cm}|p{0.8cm}|p{0.8cm}|p{0.8cm}|p{0.8cm}|p{0.8cm}|p{0.8cm}|p{0.8cm}|p{0.8cm}|p{0.8cm}|} \hline
\textbf{Question} & \textbf{Ground Truth} & \textbf{Claude Instant} & \textbf{Claude v2.1} & \textbf{Claude v2} & \textbf{Claude v3 Sonnet} & \textbf{Claude v3 Haiku} & \textbf{Claude v3 Opus} & \textbf{v3.5 Sonnet v1} & \textbf{v3.5 Sonnet v2} & \textbf{v3.7 Sonnet} & \textbf{v3.5 Haiku} \\ \hline
January network deductible (self) & 250.00 & 250.00 & 250.00 & 250.00 & 250.00 & 250.00 & 250.00 & 250.00 & 250.00 & 250.00 & 250.00 \\ \hline
February out-of-pocket max (spouse) & 4,000.00 & 4,000.00 & 4,000.00 & 4,000.00 & 4,000.00 & 4,000.00 & 4,000.00 & 4,000.00 & 4,000.00 & 4,000.00 & 4,000.00 \\ \hline
March HRA contribution (partner) & 83.00 & 83.00 & No resp. & No resp. & 83.00 & 83.00 & 83.00 & 83.00 & 83.00 & 83.00 & 83.00 \\ \hline
April out-of-pocket max (family) & 4,500.00 & 6,000.00 & 4,500.00 & 4,500.00 & 4,500.00 & 4,500.00 & 4,500.00 & 4,500.00 & 4,500.00 & 4,500.00 & 4,500.00 \\ \hline
May network deductible (family) & 2,000.00 & 1,500.00 & 1,000.00 & 1,000.00 & 2,000.00 & 1,334.00 & 2,250.00 & 2,000.00 & 2,000.00 & 2,000.00 & 1,134.00 \\ \hline
June out-of-pocket max (self) & 1,168.00 & 1,168.00 & No resp. & No resp. & 1,168.00 & 1,168.00 & 1,168.00 & 1,168.00 & 1,168.00 & 1,168.00 & 1,168.00 \\ \hline
July HRA contribution (self) & 250.00 & 250.00 & 250.00 & 250.00 & 250.00 & 250.00 & 250.00 & 250.00 & 250.00 & 250.00 & 250.00 \\ \hline
August network deductible (child) & 834.00 & 666.00 & No resp. & 834.00 & 834.00 & 834.00 & 834.00 & 834.00 & 834.00 & 834.00 & 834.00 \\ \hline
September deductible (family) & 1,000.00 & 1,000.00 & 1,000.00 & 1,000.00 & 1,000.00 & 1,000.00 & 1,000.00 & 1,000.00 & 1,000.00 & 1,000.00 & 1,000.00 \\ \hline
October out-of-pocket max (partner) & 1,000.00 & 1,000.00 & 1,000.00 & 1,000.00 & 1,000.00 & 1,000.00 & 1,000.00 & 1,000.00 & 1,000.00 & 1,000.00 & 1,668.00 \\ \hline
\textbf{Accuracy} & - & 70\% & 60\% & 70\% & 100\% & 90\% & 90\% & 100\% & 100\% & 100\% & 80\% \\ \hline
\textbf{No-Response Rate} & - & 0\% & 30\% & 20\% & 0\% & 0\% & 0\% & 0\% & 0\% & 0\% & 0\% \\ \hline 
\end{tabular} \caption{Detailed Performance Comparison of Claude Models for question and answering on HR Benefit Questions (sampled from test-dataset). \textit{Note: All monetary values have been masked and replaced with simulated data.}}
\label{tab:detailed_model_comparison}
\end{table*}

The data reveals several important patterns in model performance. A clear progression in capability is evident from earlier Claude models (Instant, v2, v2.1) to the Claude v3 family and beyond, with Claude v3 Sonnet achieving perfect accuracy on our test set. Earlier models (Claude v2, v2.1) completely failed to retrieve relevant information for 20-30\% of queries, returning "No response found" for questions about specific benefit categories. When providing incorrect answers, these models typically struggled with confusing benefit tiers (e.g., providing family benefits when asked about individual coverage), misinterpreting temporal information (providing incorrect monthly figures), and inaccurately processing numerical values in complex table structures. In contrast, Claude v3 Sonnet, v3.5 Sonnet (v1 \& v2), and v3.7 Sonnet achieved perfect accuracy across all queries, while Claude v3 Haiku and Claude v3 Opus demonstrated strong but not perfect performance (90\% accuracy). This detailed analysis supports our selection of Claude v3 models for HR document processing tasks, with Claude v3 Sonnet offering the optimal balance of accuracy and efficiency for production deployment in talent management applications.

\section{System Prompt} \label{appendix_prompt}
This section provides the prompt used to convert the tables embedded in images to plain text, where the extracted text is directly used in RAG retrieval activities. 
\begin{verbatim}
def multi_modality_enhancement_prompt(image):
    prompt = f"""###Instruction: Act as an intelligent 
    document processing agent specialized in 
    talent management documentation. Your task is to:
    1. Parse the hierarchical table embedded in image
    2. Convert each table cell into contextually meaningful 
       text while preserving:
       - Row-column relationships and Numerical precision
       - Contextual dependencies
    3. Generate complete, well-formed sentences that capture:
       - Cell value and its position context
       - Related header information
       - Any conditional relationships
    For the given HR document image {image}, maintain data 
    fidelity while ensuring the output is optimized 
    for downstream RAG chat applications to support user's
    question-anwering in real-time.
    ###Response: """
    return prompt
\end{verbatim}
\label{fig:custom_prompt}

\end{document}